\documentclass[runningheads]{llncs}
\usepackage{graphicx}
\begin{document}
\title{From Knowledge Representation  to \\ Knowledge Organization and Back}
\titlerunning{From KR to KO and Back}
% If the paper title is too long for the running head, you can set
% an abbreviated paper title here

\author{Fausto Giunchiglia \orcidID{0000-0002-5903-6150} \and
Mayukh Bagchi \orcidID{0000-0002-2946-5018}}
% %
\authorrunning{Fausto Giunchiglia and Mayukh Bagchi}

\institute{DISI, University of Trento, Italy. \\
\email{\{fausto.giunchiglia,mayukh.bagchi\}@unitn.it}\\}
\maketitle              % typeset the header of the contribution
\begin{abstract}
Knowledge Representation (KR) and facet-analytical Knowledge Organization (KO) have been the two most prominent methodologies of data and knowledge modelling in the Artificial Intelligence community and the Information Science community, respectively. KR boasts of a robust and scalable ecosystem of technologies to support knowledge modelling while, often, underemphasizing the quality of its models (and model-based data). KO, on the other hand, is less technology-driven but has developed a robust framework of guiding principles (\textit{canons}) for ensuring modelling (and model-based data) quality. This paper elucidates both the KR and facet-analytical KO methodologies in detail and provides a functional mapping between them. Out of the mapping, the paper proposes an integrated KO-enriched KR methodology with all the standard components of a KR methodology \textit{plus} the guiding canons of modelling quality provided by KO. The practical benefits of the methodological integration has been exemplified through a prominent case study of KR-based image annotation exercise. 

\end{abstract}

\section{Introduction}
\label{S1}
% \begin{itemize}
%     \item Paragraph (1): Contextualization of KR 
%     \item Paragraph (2): Contextualization of KO
%     \item Paragraph (3): Problem - Data Quality
%     \item Paragraph (4): Solution: (1) Mapping between KR and KO processes, (2) Enriched KR Methodology
%     \item Paragraph (5): Paper Organization.
% \end{itemize}

% Paragraph (1): Contextualization of KR 
Knowledge Representation is the arena of Artificial Intelligence (AI) dealing with \emph{``how knowledge can be represented symbolically"} \cite{2004-KR} within intelligent systems. To that end, KR encompasses a wide spectrum of advanced \emph{technologies} (e.g., the Semantic Web technology stack \cite{2009-SWTS}) and \emph{methodologies} (e.g., \cite{1997-methontology,2014-KO}) to facilitate generation of KR artifacts (e.g., ontologies \cite{2009-ontology}, conceptual models \cite{2014-OAP}, Knowledge Graphs (KGs) \cite{2021-KG}). Such methodologies have been widely adopted or adapted for application scenarios from as conventional as, e.g., data integration \cite{2020-SDI}, %wherein, the focus is to exploit KR to integrate data coming from heterogeneous data silos,%
to as innovative as, e.g., image annotation \cite{2023-iconf}, wherein, conceptual models are designed to organize (objects in) images which are later exploited as high-quality training data for computer vision tasks like object recognition. However, while KR methodologies have remained highly mature in terms of supporting technology and scalable in terms of technology-enabled services, a \emph{key} criticism has been that they have traditionally underemphasized \emph{modelling quality}, e.g., of ontologies or conceptual models (see, \cite{2005-FLAIRS,2013-QQUARE,2015-OAP}, for a few prominent studies), resulting in, often, flawed and biased datasets designed according to such models.

% Paragraph (2): Contextualization of KO
Knowledge Organization, on the other hand, is the arena of Information Science dealing with the cumulative set of activities concerning, quoting \cite{2008-KO}, the \emph{``description, indexing and classification"}  of information resources (e.g., books) in different kinds of \textit{`memory institutions'} (e.g., libraries). To that end, KO encompasses a wide spectrum of modelling systems \cite{2008-KOS}, e.g., classification schemes, taxonomies, catalogs, etc., and, different approaches \cite{2008-KO}, e.g., enumerative approach, facet-analytical approach, etc., which integrate different KO systems. In this paper, we concentrate exclusively on the facet-analytical KO approach originally proposed by Ranganathan \cite{SRR-64,SRR-67}. %For example, to develop the classification hierarchy for a subject, facet-analysis prescribes \textit{analysis}, i.e., analyzing and breaking down the different dimensions of the subject into hierarchical models of primitive atomic concepts (termed \textit{facets}), followed by \textit{synthesis}, wherein, different facets can be combined in a specific order to obtain, e.g., the final unique classification number of a book of that subject. Further, the classification number is reverse-engineered to generate the subject headings which, with other bibliographic attributes, form the description of the book within the catalogue record which is used by a library user.%
Noticeably, while the activities in (faceted) KO are predominantly intellectual in nature and less technology-driven, they have, in order to support the aforementioned activities, developed a huge number of \emph{guiding principles} for (conceptual) modelling, termed \emph{canons}, following which \emph{high-quality models}, .e.g., conceptual hierarchies, conceptual models and model-based \emph{datasets}, can be generated. 

% Paragraph (3): Problem: Data Quality
From the above overviews, we notice that both KR and KO have several complementary strengths and weaknesses. We concentrate on the specific \textit{problem of how the modelling quality of KR artifacts can be methodologically ensured}. One clear way, in the light of the above discussion, can be the incorporation of the guiding principles for modelling, namely, the \emph{canons}, developed within the facet-analytical approach of KO in mainstream KR. In fact, the potentiality of such an incorporation can be extremely crucial because of growing concerns about \textit{systematic design flaws} inherent in the \emph{quality of data} which are designed using KR models, e.g., data quality issues \cite{2020-ICML} in ImageNet \cite{2009-imagenet}, and, additionally, the \emph{decline in both the quality and performance} of resulting data-driven models, e.g., \cite{2021-CHI,2021-NIPS}. In this paper, we propose a KO-enriched KR methodology which not only includes the standard roles and activities key to the generation of a KR model, e.g., an ontology-driven KG, but also, most importantly, incorporates the otherwise \textit{missing} activity of generating the high-quality ontology model structuring the KG following the guiding \textit{canons} prescribed in the facet-analytical approach of KO. It also presents brief highlights of a recent study-cum-experiment in KR-based image annotation \cite{ECAI23} which preliminarily validates the benefits of implementing an adapted version of the aforementioned methodology. 

The remainder of the paper is organized as follows: Section \ref{S3} and Section \ref{S4} describes the roles and activities within the different phases of a standard KR and facet-analytical KO methodology, respectively. Section \ref{S5} maps and integrates the methodologies in the previous two sections into a single KO-enriched KR methodology, with special emphasis on the activity which exploits the guiding principles (canons) to ensure model (and model-based data) quality. Section \ref{S6} describes a case study in image annotation which preliminarily validates the advantages of the integrated KO-enriched KR methodology. Section \ref{S7} discusses related work and Section \ref{S8} concludes the paper.

\section{Knowledge Representation (KR)}
\label{S3}
% \begin{itemize}
%     \item Paragraph (1): Section Introduction + Describe Fig.
%     \item Paragraph (2): ETG Development (Upper Half of Fig-3)
%     \item Paragraph (3): Ontology Rep Development (Lower Half of Fig-3)
%     \item Paragraph (4): KG Engineering (Merging process of Fig-3)
%     \item Paragraph (5): KG User - Use and Reuse - Activities (Role - 3 of Fig-3)
%     \item Paragraph (6): Observations/Disadvantages
% \end{itemize}

\begin{figure}[htp]
%\vspace{-0.6cm}
\includegraphics[width=12cm,height=8cm]{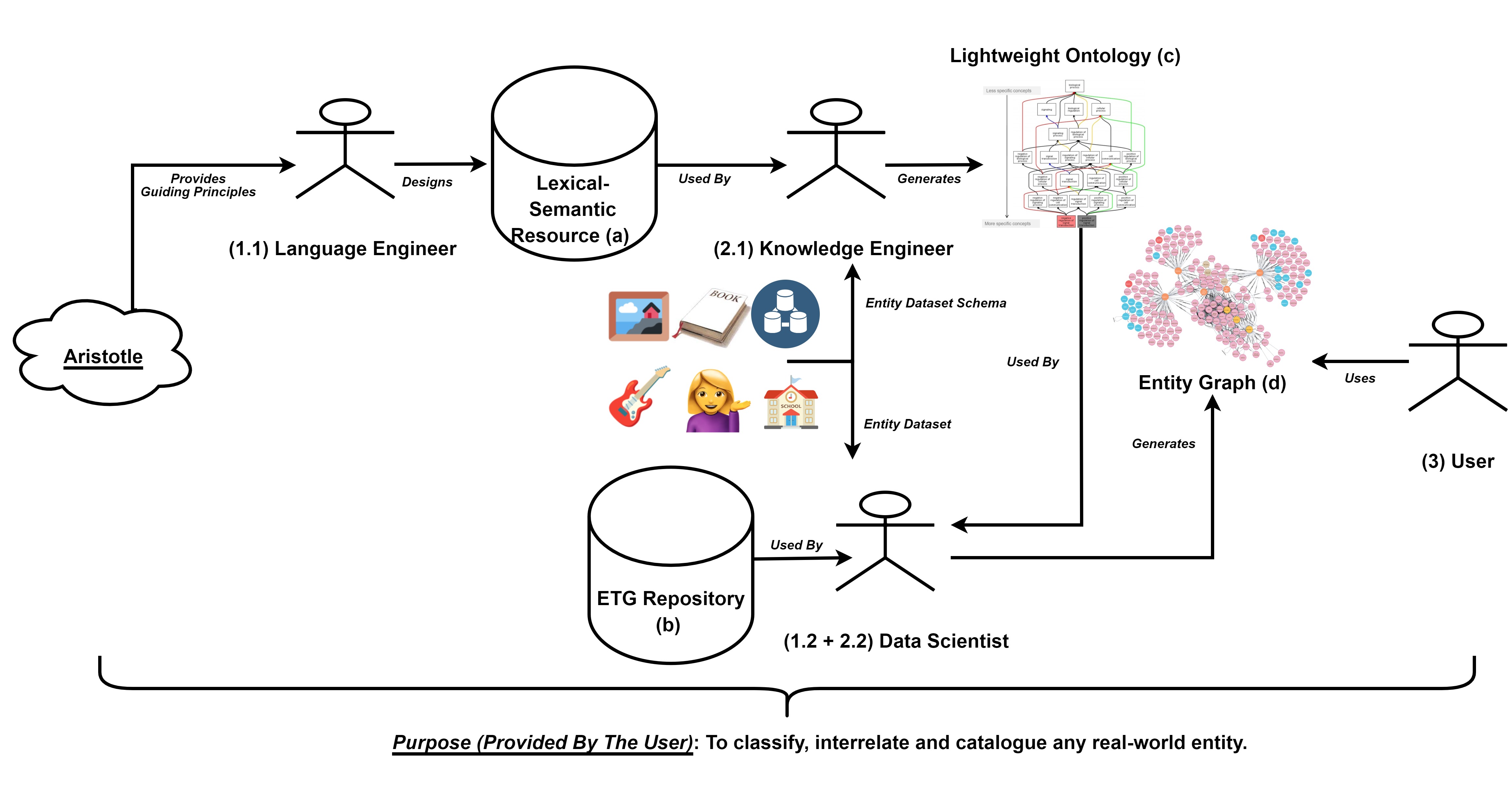}
\caption{A high-level view of the KR Methodology.}
\centering
\label{I1}
\end{figure}

%  Paragraph (1): Section Introduction + Describe Fig
\noindent Let us first concentrate on a detailed exposition of the various phases, and within each phase, the various roles, activities and artifacts, which together compose to form a standard KR methodology. See Figure \ref{I1} for a high-level view of the KR methodology. The methodology can be seen as being constituted of the following four distinct phases (the terminologies being detailed later):
\begin{enumerate}
    \item The first phase initiating with guiding principles and concluding with the generation of the Lightweight Ontology.
    \item The second phase concerning the development of the Entity Type Graph (ETG) repository.
    \item The third phase of the methodology concentrating on how the data scientist takes in three different inputs, namely, the Lightweight Ontology, Entity Dataset(s) and ETGs from the ETG repository, and, suitably integrates them to generate the Entity Graph (EG).
    \item Finally, the fourth and the final phase concentrating on the different ways in which a user can use and exploit the EG.
\end{enumerate}
The diagrammatic symbols of Figure \ref{I1} include, amongst others, the various roles (visualized via \textit{actor} icon), activities (visualized as edge \textit{labels}) and artifacts (visualized variously via other (coloured) icons). Notice also that the numbers and lowercase alphabets which identify roles and artifacts, respectively, in Figure \ref{I1}, are employed later (in Section \ref{S5}) for functional mapping purposes. We now consider each phase of the methodology individually.

%  Paragraph (2): ETG Development (Upper Half of Fig-3)
The first phase, as already noted, commences with guiding principles articulated as the intensional definition-building paradigm of \textit{Genus-Differentia} \cite{GD} proposed by Aristotle (visualized via a cloud) over two millennia ago. According to the paradigm, the linguistic definition of any real-world object \cite{2017-OAF} expressed, e.g., as a noun, is formulated in terms of two constituent definitions: \textit{Genus} and \textit{Differentia}. While \textit{Genus} defines an \emph{a priori} set of properties shared across distinct objects, e.g., the property of being a stringed musical instrument, \textit{Differentia} defines a novel set of properties used to differentiate objects having the same \textit{Genus}, e.g., the properties of musical instruments having six strings or thirteen strings. Therefore, as illustrated in Figure \ref{I1}, the \textit{Genus-Differentia} guidelines are taken in input and adhered to by the \textit{Language Engineer} to create machine-processable language data, e.g., WordNet-like lexical-semantic hierarchies of synsets \cite{PWN} codifying word meanings in different natural languages and in different domains \cite{2004-WDH} represented in a machine processable format, e.g., Lexical Markup Framework \cite{2014-LMF}. 

Such machine-processable language data are stored and managed as part of a \textit{Lexical-Semantic Resource} (see Figure \ref{I1}) which is usually designed as a collection of WordNet-like machine-processable lexical hierarchies and, in some cases \cite{UKC-IJCAI,UKC-CICLING}, with an additional language-independent semantic layer unifying different language-specific lexical hierarchies. Given the design of the \textit{Lexical-Semantic Resource}, the final activity of this phase shifts to the \textit{Knowledge Engineer} (see Figure \ref{I1}) who has to now generate the \textit{Lightweight Ontology} (Figure \ref{I1}) which is an intermediate machine-processable formal hierarchy \emph{``consisting of backbone taxonomies"} \cite{2008-LWO} that are being considered for representing knowledge in the context of a specific purpose provided by the user. To that end, the \textit{Knowledge Engineer} has to take in two important inputs. Firstly, (s)he has to take in input the appropriate lexical-semantic hierarchy of words in a specific language which will inform the syntax and modelling of the taxonomical hierarchy of the \textit{Lightweight Ontology} she will generate. In addition, (s)he also takes in input the (dataset) \textit{schema} of the entities which (s)he wants to model in the \textit{Lightweight Ontology}, this, providing her with the \textit{exact} way in which the subsumption hierarchy of parent and child concepts, pertaining to the entities (and their datasets), should be organized\footnote{thereby, significantly speeding up the process of merging the datasets with the KR model at a later stage of the methodology.}.

%  Paragraph (3): Ontology Rep Development (Lower Half of Fig-3)
The second phase, as illustrated in Figure \ref{I1}, concerns the design and development of a repository (exposed via a catalog) of reusable Entity Type Graphs (ETGs) \cite{2023-LS}, wherein, ETGs are defined as machine-processable \textit{ontological} representations formalizing entity types \cite{ETR} which capture the semantics inherent in (dataset) entities. Notice the fact that, ETGs, being \textit{ontological} representations, also encode \textit{object properties} (modelling how an entity is related to other entities) and \textit{data properties} (modelling the attributes which describe an entity) as is standard to any KR model. This repository is crucial to the methodology in all the four dimensions - \textit{F}indability, \textit{A}ccessibility, \textit{I}nteroperability, and \textit{R}eusability - advanced by the \textit{FAIR} paradigm \cite{fair} of scientific data management. Especially, the repository would facilitate not only interoperability amongst ETGs (as they are modelled following the same technological standards) but also enable ETG-based interoperability of data (when ETGs from the repository are reused, in conjunction with lightweight ontology, to produce the final EG). It would also promote the circular reuse of ETGs, for instance, the reuse of the same ETG but for different use case scenarios of KR. A crucial observation. Notice that while there are several technical advantages of designing an ETG repository as elucidated above, the methodology \textit{does not} specify any activity path or roles for enforcing guiding principles which ensure the high-quality of the ETG models constituting the repository (noticeable by the lack of any activity preceding the ETG repository in Figure \ref{I1}).

% Paragraph (4): KG Engineering (Merging process of Fig-3)
The third phase concentrates on how the \textit{Data Scientist} (see Figure \ref{I1}) exploits the outputs of the previous two phases and the concrete (entity) datasets of the use case at hand to generate the final Entity Graph (EG). An EG is a type of Knowledge Graph (KG) which is: (i) taxonomically structured via a lightweight ontology, and, (ii) interrelated and described with object properties and data properties from the relevant ETG. To that end, the \textit{Data Scientist} receives in input the lightweight ontology (output of the first phase), wherein, the use-case specific concepts are hierarchically modelled following the appropriate lexical-semantic hierarchy. The lightweight ontology is then grounded in the relevant classificatorily-compliant ETG from the ETG repository (output of the second phase). This activity has three key advantages. Firstly, the ETG endows the lightweight ontology-ETG combined KR artifact with crucial object properties (which interrelate, at a conceptual level, the entities it encode) and data properties (which encode the descriptive attributes of entities). Secondly, the grounding also ensures the completeness of the KR model in terms of grounding use-case specific concepts (modelled bottom-up) into general (domain) concepts (modelled top-down). Thirdly, the above grounding also ensures the fact that, later, the data encoded via the specific lightweight ontology becomes interoperable with data encoded via other lightweight ontologies which commit to the same ETG. Given the lightweight ontology-ETG combined KR artifact, the \textit{Data Scientist} takes in input the entity datasets and semi-automatically maps them to the combined artifact (see, \cite{KGSWC}, for the detailed process), one dataset at a time, to generate the final Entity Graph (EG) which is a KG modelled by populating the nodes of the combined KR artifact with data.

% Paragraph (5): KG User - Use and Reuse - Activities (Role - 3 of Fig-3)
Finally, the fourth phase of the methodology concentrates on the different ways in which different \textit{users} can exploit the Entity Graph (EG). This, in turn, informs as well as depends on the different type of basic and specialized \textit{services} which can be developed to explore the EG. A set of basic services can include the option to download a (part of a) KG in different formats, the option to query a KG (see, \cite{2023-LS}, for an enumeration of potential services). Advanced services can be designed depending upon the use-case, for instance, to exploit a generated KG for image annotation tasks. A prime example can be of an interactive service to facilitate \textit{egocentric} image annotation (with egocentrism becoming increasingly pivotal to the computer vision community \cite{2020-CEOR}). Such a service, for instance, would allow a user to use (a portion of) a KG schema and populate each of its node (i.e., a concept) with (multiple) relevant images to generate a hierarchically organized image dataset ready to be downloaded and used as training data for computer vision models.

%  Paragraph (6): Observations/Disadvantages
% There are a few important highlights to note with respect to the above KR methodology (Figure \ref{I1}). Firstly, notice that the methodology is completely \textit{general} in its scope and functionality as it can completely accommodate any level and form of \textit{knowledge diversity} \cite{2006-ECAI}, e.g., in terms of languages, ontological models, domains and entities. Secondly, the \textit{purpose} behind a particular use-case of the methodology is always provided \textit{by the user} and it is always the purpose which determines the selection, classification, interrelation and descriptions of entities modelled via the methodology. Thirdly, and perhaps the most important, notice, as also depicted in Figure \ref{I1}, the methodological flow of roles and activities prior to the ETG repository is \textit{completely blank}. In fact, this blank segment of the methodology should \textit{otherwise} have been the most consequential part of the methodology determining and ensuring the classification-intensive \textit{modelling quality} of the ETGs, and thereby, of the final output KR model, i.e., the EG.

\section{Knowledge Organization (KO)}
\label{S4}
% \begin{itemize}
%     \item Paragraph (1): Section Introduction + Describe Fig.
%     \item Paragraph (2): Faceted Classification (Upper Half of Fig-2)
%     \item Paragraph (3): Cataloguing Code (Lower Half of Fig-2 connected to Upper Half)
%     \item Paragraph (4): Cataloguing process (Merging process of Fig-2)
%     \item Paragraph (5): Library User Activities (Role - 3 of Fig-2)
% \end{itemize}

\begin{figure}[htp]
%\vspace{-0.6cm}
\includegraphics[width=12cm,height=8cm]{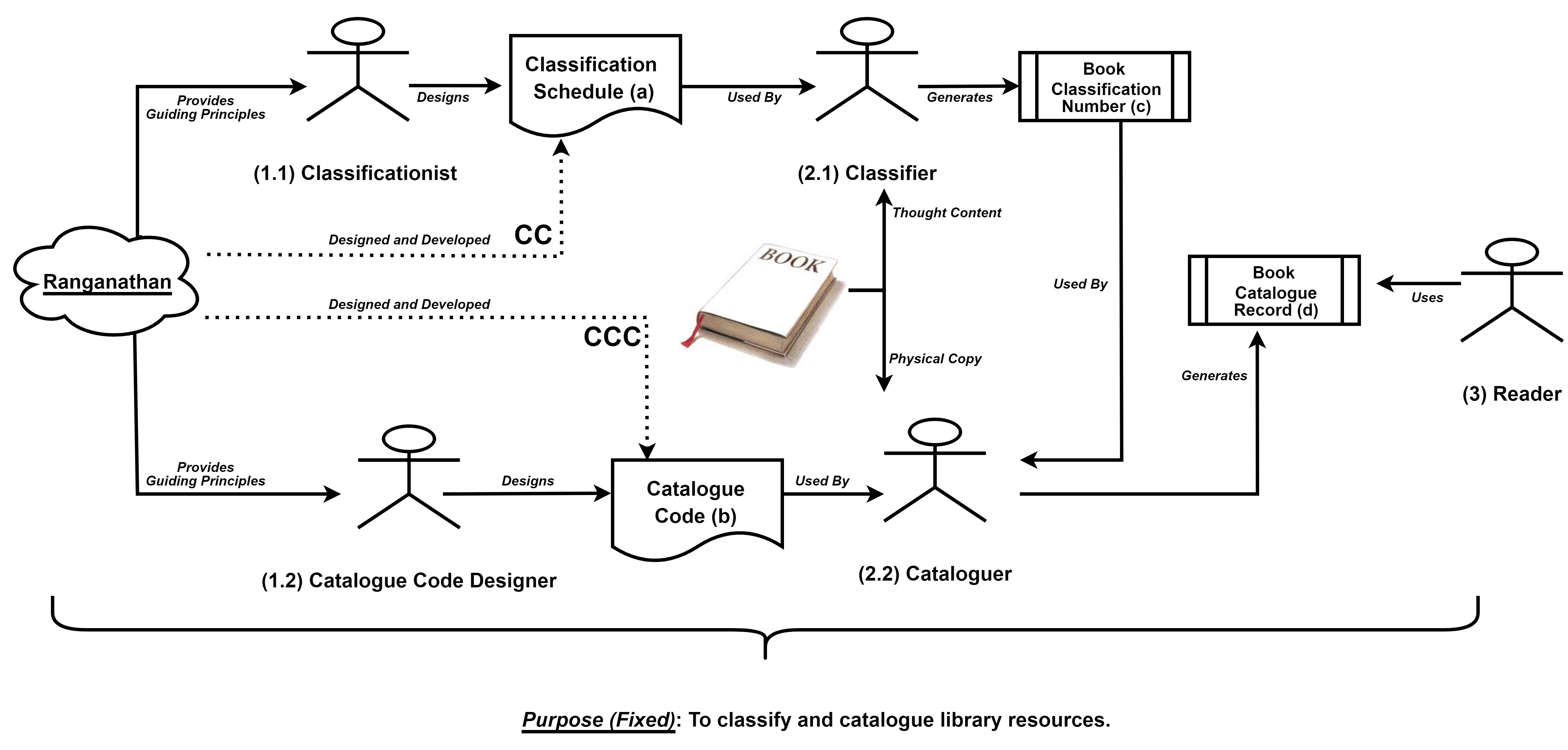}
\caption{A high-level view of the facet-analytical KO Methodology.}
\centering
\label{I2}
\end{figure}

% Paragraph (1): Section Introduction + Describe Fig.
\noindent In sync with the previous section on the KR methodology, let us now concentrate on a detailed exposition of the various phases which together compose to form a standard KO methodology within the facet-analytic tradition of KO. See Figure \ref{I2} for a high-level view of the KO methodology. The methodology can be seen as being constituted of the following four distinct phases (the terminologies being detailed later):
\begin{enumerate}
    \item The first phase initiating with the guiding principles for \emph{Classification} \cite{SRR-67} and concluding with the generation of the book classification number.
    \item The second phase concerned with the guiding principles for \emph{Cataloguing} \cite{SRR-64} and concluding with the design of the catalogue code.
    \item The third phase of the methodology concentrating on how the cataloguer takes in three different inputs, namely, the book classification number, the physical copy of the book and the catalogue code, and, suitably integrates them to generate the book catalogue record. 
    \item Finally, the fourth and the final phase concentrating on the different ways in which a reader visiting a library can use and exploit the book catalogue record.
\end{enumerate} 
The diagrammatic symbols of Figure \ref{I2} are similar as before. We now consider each phase of the methodology individually.

% Paragraph (2): Faceted Classification (Upper Half of Fig-2)
The first phase, as already noted, commences with guiding principles, termed \textit{canons of classification}, postulated by Ranaganathan \cite{SRR-67} to oversee the generation of \textit{high-quality} classification hierarchies for any subject. Notice that, by \textit{subject}, Ranganathan meant not only the universe of \emph{macro thought} (e.g., disciplines such as Mathematics, Chemistry) but also, crucially, the universe of \textit{micro thought} \cite{SRR1954} (e.g., depth classification of differential equations). We now enumerate below an overview of the three groups of \textit{canons of classification} (detailed in \cite{SRR-67}) which are core to the methodology illustrated in Figure \ref{I2}:
\begin{enumerate}
    \item \textit{Canons of Idea Plane}, which focus on the modelling of concepts in a classification hierarchy based on their perceivable properties. They include:
    \begin{enumerate}
        \item canons about \textit{characteristics} based on which concepts should be differentiated with respect to a single level in the classification hierarchy, e.g., canon of \textit{relevance}, canon of \textit{ascertainability}.
        \item canons about \textit{succession of characteristics}, i.e., how differentiating characteristics should succeed one after the other with respect to a classification hierarchy, e.g., canon of \textit{relevant succession}.
        \item canons about \textit{arrays} based on which concepts, within the same horizontal level in the classification hierarchy, should be modelled, e.g., canon of \textit{exhaustiveness}.
        \item canons about \textit{chains} based on which concepts, within a single hierarchical path in the classification hierarchy, should be modelled, e.g., canon of \textit{modulation}.
    \end{enumerate}
   \item \textit{Canons of Verbal Plane}, which focus on the proper linguistic rendering of the concepts modelled following the canons of the Idea Plane, e.g., canon of \textit{reticence}.
   \item \textit{Canons of Notational Plane}, which focus on assigning a unique numerical identifier for each linguistically labelled concept in the classification hierarchy, e.g., canon of \textit{synonym} and canon of \textit{homonym}. 
\end{enumerate}

\noindent Thereafter, as illustrated in Figure \ref{I2}, the \textit{canons of classification} as briefed above are taken in input by the \textit{Classificationist} to design (a set of) faceted classification schedules for either general knowledge organization usage, e.g., the Colon Classification (CC) \cite{CC} designed and developed by Ranganathan himself (indicated with dashed lines), or, for specialized knowledge organization usage, e.g., Uniclass\footnote{https://www.thenbs.com/our-tools/uniclass}, specialized classification for the construction sector. Given the design of the classification schedule(s), the final activity of this phase shifts to the \textit{Classifier} (see Figure \ref{I2}) who has to now generate the unique \textit{Book Classification Number} (Figure \ref{I2}), e.g., Colon Number, for the subject matter of a book. To that end, the \textit{Classifier} has to take in two important inputs. Firstly, (s)he has to take the \emph{a priori} designed classification schedules which provide him/her with the concept hierarchy (with each concept uniquely identified via an identifier) as well as the formula (i.e., the \textit{facet formula}) in which relevant concepts should be combined to generate the book classification number. Secondly, (s)he has to take in input the \textit{thought content} of the book to be classified. Finally, the \textit{Classifier} follows Ranganathan's analytico-synthetic classification number generation procedure \cite{CC} to generate the \textit{Book Classification Number} which uniquely identifies its subject matter. Notice that the above process is valid not only for books but also for any library resource.

%  Paragraph (3): Cataloguing Code (Lower Half of Fig-2 connected to Upper Half)
The second phase, as illustrated in Figure \ref{I2}, concerns the design and development of a catalogue code. It commences with the \textit{canons of cataloguing} as guiding principles postulated by Ranaganathan \cite{SRR-64} to oversee the generation of \textit{high-quality} description of any book (or, any library resource). Some of the canons are notable to be briefed. For example, the canon of \textit{sought heading} mandates that the metadata attributes which should be captured about a book in a catalogue should be strictly based on the \textit{likelihood} of how a user might approach the catalogue. To that end, all unnecessary metadata should be excluded from the catalogue record. Further, the canon of \textit{consistence} prescribes that, for a specific type of (library) resource, the set of metadata which constitute its catalogue record should be consistent, unless otherwise prescribed by the canon of \textit{context}. Another crucial principle is that of \textit{local variation} which allows flexibility in the description of a catalogue record if need arises due to a very typical resource specific to a context. These canons of cataloguing, amongst many others detailed in \cite{SRR-64}, are taken as input guidelines by the \textit{Catalogue Code Designer} to finally design a \textit{Catalogue Code} - a body of specifications on how and what metadata should be encoded in a catalogue record for a specific bibliographic resource type. In fact, Ranganathan himself designed and developed one such catalogue code termed the \textit{Classified Catalogue Code} (or, CCC; shown in Figure \ref{I2} via dashed lines).

%  Paragraph (4): Cataloguing process (Merging process of Fig-2)
The third phase concentrates on how the \textit{Cataloguer} (see Figure \ref{I2}) exploits the outputs of the previous two phases and the concrete physical copy of the resource (e.g., book) at hand to generate the final \textit{Book Catalogue Record}. A bibliographic catalogue record encodes metadata which are termed as \textit{access points}, e.g., title, author, year of publication, subject headings, etc., which might be help the reader in her quest to search and identify the actual copy of the book (or, resource). To that end, the \textit{Cataloguer} receives in input the book classification number (output of the first phase), wherein, the subject matter of the book is modelled following the appropriate classification schedule (and facet formula). Given the book classification number, the procedure of \textit{Chain Indexing} is performed, whereby, the classification number is reverse-engineered through a series of specified steps (see \cite{SRR-64} for details) to generate subject headings (\textit{tags} in modern parlance) which can serve as access points in the catalogue record. Further, the \textit{Cataloguer} also receives two other inputs: the concrete copy of the book which contains all its imprint details, and, the catalogue code (e.g., CCC) which strictly specifies which and how such imprint information should be modelled in the catalogue record. Thereafter, the \textit{Cataloguer} integrates the subject headings together with the requisite imprint attributes and the classification number (together with the \textit{call number} to identify the book's exact place in the shelves) to generate the final \textit{Book Catalogue Record}.

% Paragraph (5): Library User Activities (Role - 3 of Fig-2)
Finally, the fourth phase concentrates on how a library \textit{reader} can exploit the book catalogue record. In addition to the usual ways of using the catalogue (see, e.g., \cite{SRR-64}), Ranganathan's \texttt{APUPA} principle \cite{DACC} facilitates a reader in finding very related and somewhat related books/resources on either side of the particular resource one is searching for). Further, a reader can use the catalogue record of an Online Public Access Catalog (OPAC) which is enhanced with \textit{library discovery services} \cite{LRD}, thereby, going beyond traditional means of library search to include web-scale \textit{exploratory search} and recommendations.

%  Paragraph (6): Observations/advantages
% There are a few observations with respect to the above facet-analytical KO methodology (Figure \ref{I2}). Firstly, notice that the methodology is \textit{specialized} in its scope and functionality as it can accommodate any level and form of resources in the \textit{bibliographic} universe, e.g., in terms of classifications, cataloguing descriptions, subjects and bibliographic entities (e.g., books). Secondly, the \textit{purpose} behind a particular use-case of the methodology is always \textit{fixed}, namely, the methodology is specialized and limited to the classification and cataloguing of bibliographic entities. Last but not the least, as also depicted in Figure \ref{I2}, the methodological flow of roles and activities prior to both the classification schedule design and the catalogue code design in \textit{highly robust} in terms of adherence to supporting guiding principles. In fact, the guiding principles (canons) for both classification and description (i.e., cataloguing) incorporated in the facet-analytical KO methodology decisively ensures the \textit{modelling quality} of the classification hierarchies, and thereby, of the final output KO model, i.e., the catalogue record.

\section{From KR to KO and Back}
\label{S5}

\begin{table}[]
\resizebox{\textwidth}{!}{%
\begin{tabular}{|l|l|l|l|}
\hline
\textbf{Phase} &
  \textbf{KR} &
  \textbf{KO} &
  \textbf{KO-Enriched KR} \\ \hline
1 &
  \begin{tabular}[c]{@{}l@{}}Language Engineer\\ (1.1)\end{tabular} &
  \begin{tabular}[c]{@{}l@{}}Classificationist\\ (1.1)\end{tabular} &
  \begin{tabular}[c]{@{}l@{}}Language Engineer\\ (1.1)\end{tabular} \\ \hline
2 &
  \begin{tabular}[c]{@{}l@{}}Knowledge Engineer\\ (2.1)\end{tabular} &
  \begin{tabular}[c]{@{}l@{}}Classifier\\ (2.1)\end{tabular} &
  \begin{tabular}[c]{@{}l@{}}Knowledge Engineer\\ (2.1)\end{tabular} \\ \hline
3 &
  \begin{tabular}[c]{@{}l@{}}Data Scientist\\ (1.2 + 2.2)\end{tabular} &
  \begin{tabular}[c]{@{}l@{}}Catalogue Code Designer\\ (1.2)\\ \\ Cataloguer\\ (2.2)\end{tabular} &
  \begin{tabular}[c]{@{}l@{}}Ontology Engineer\\ (1.2 + 2.3)\\ \\ Data Scientist\\ (2.2)\end{tabular} \\ \hline
4 &
  \begin{tabular}[c]{@{}l@{}}User\\ (3)\end{tabular} &
  \begin{tabular}[c]{@{}l@{}}Reader\\ (3)\end{tabular} &
  \begin{tabular}[c]{@{}l@{}}User\\ (3)\end{tabular} \\ \hline
\end{tabular}%
}
\\
\caption{Functional mapping between roles of KR, KO and KO-Enriched KR.}
\label{T1}
\end{table}

\noindent In the previous two sections, we've provided a detailed elucidation of the principle roles, activities and artifacts of both the Knowledge Representation and the facet-analytical Knowledge Organization methodology, respectively. In this section, before showing how to complete the loop \textit{from KR to KO and back}, we first concentrate on a \textit{functional mapping} of, chiefly, the roles and the artifacts, of the two aforementioned methodologies (see the first three columns of Table \ref{T1}). Notice two things. Firstly, via functional mapping, we ascertain whether or not two roles or artifacts perform the \textit{same or synonymous} broad function, irrespective of differences in their syntax, semantics or form. Secondly, in Table \ref{T1}, we illustrate only the roles as they embody the major functional differences between the two methodologies and not the artifacts (duely elucidated in the following description) which are functionally synonymous. We constantly refer to Table \ref{T1}, and, to Figure \ref{I1} and Figure \ref{I2} as required, in the following discussion.

At the very outset, note that we proceed the functional mapping on the basis of the four informal phases via which we detailed each individual methodology. We notice that, in the first phase, there is a mapping between the roles of \textit{Language Engineer (1.1)} and that of \textit{Classificationist (1.1)}. This is clearly due to the fact that both these roles, based on input guiding principles, generate lexical-semantic classification hierarchies. Next, we also note that the artifacts that the above two roles produce, namely, the \textit{Lexical-Semantic Resource (a)} (see Figure \ref{I1}) and the \textit{Classification Schedule (a)} (see Figure \ref{I2}) are also in functional mapping to each other, given that both are essentially constituted of (a set of) lexical-semantic classification hierarchies focused on different subjects, domains, etc. Further, the roles \textit{Knowledge Engineer (2.1)} and \textit{Classifier (2.1)} are also mapped to each other because their central function is to use prescribed lexical-semantic hierarchies and classify the \textit{subject} of resources according to their relevant concept within the hierarchy. To that end, the output artifacts they produce, i.e., \textit{Lightweight Ontology (c)} (see Figure \ref{I1}) and \textit{Book Classification Number (c)} (see Figure \ref{I2}) also serve the synonymous function of encoding the categorization of either the entity dataset schema or the book subject matter.  

In the second phase, the \textit{ETG Repository (b)} (see Figure \ref{I1}) and \textit{Catalogue Code (b)} (see Figure \ref{I2}) are also functionally synonymous in the sense that both of these artifacts endow their input artifacts with reusable specifications of how to describe concepts via attributes within a classification hierarchy. The first major break in functional mapping occurs with respect to the role of \textit{Data Scientist (1.2 + 2.2)}. The Data Scientist role subsumes two roles, namely, that of the \textit{Catalogue Code Designer (1.2)} which exclusively deals with the specification of the descriptive (data) attribute schema in addition to that of the \textit{Cataloguer (2.2)} whose function is to generate the book catalogue record. Further, notice the functional synonymity between the two artifacts: \textit{Entity Graph (d)} (see Figure \ref{I1}) and \textit{Book Catalog Record (d)} (see Figure \ref{I2}), both of whose function is to provide the end user with an integrated view of top-down and bottom-up knowledge. Finally, the roles of a \textit{User (3)} and a \textit{Reader (3)} are also mapped as both function as end-users using the respective output artifacts of their methodologies in different ways. Notice also the fact that the activities across both the methodologies are also functionally mapped (see the labelled edges on Figure \ref{I1} and Figure \ref{I2}, respectively) with the only exception of the gap prior to the ETG repository in Figure \ref{I1}.

\begin{figure}[htp]
%\vspace{-0.6cm}
\includegraphics[width=12cm,height=8cm]{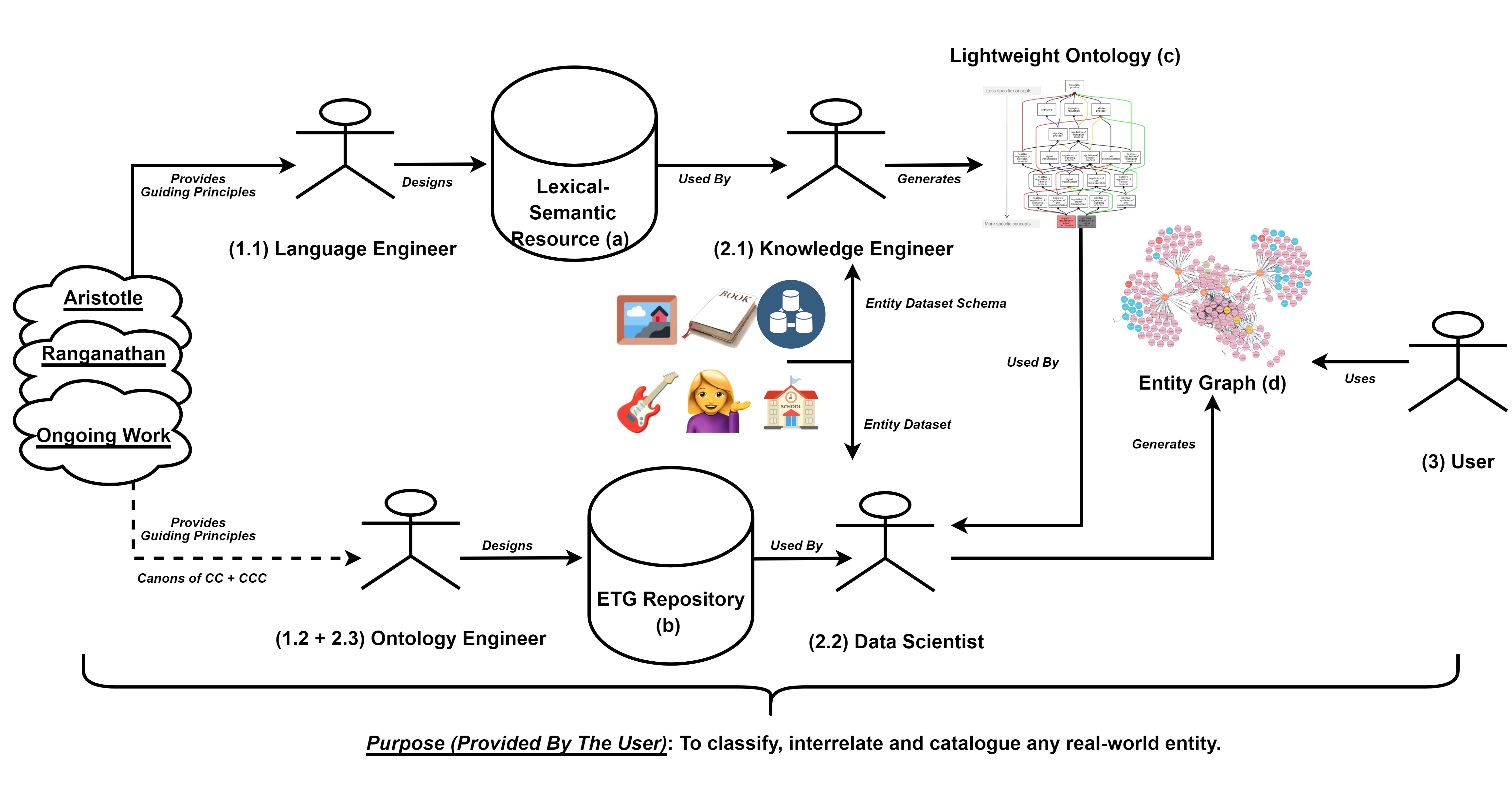}
\caption{A high-level view of the KO-Enriched KR Methodology.}
\centering
\label{I3}
\end{figure}

\noindent Given the functional mapping, we now concentrate on how the guiding principles, i.e., \textit{canons}, advanced by the facet-analytic KO approach can be incorporated within the KR methodology to ultimately result in a KO-enriched KR methodology. To that end, we concentrate on the various phases (harmonized from both the KR and KO methodologies via the functional mapping) which together compose to form the KO-enriched KR methodology (see Figure \ref{I3}). The methodology can be seen as being constituted of the following four distinct phases:
\begin{enumerate}
    \item The first phase initiating with guiding principles and concluding with the generation of the Lightweight Ontology.
    \item The second phase initiating with guiding principles (\textit{canons}) guiding the development of \textit{high-quality} Entity Type Graphs (ETGs) within the ETG repository.
    \item The third phase of the methodology concentrating on how the data scientist generates the Entity Graph (EG), and,
    \item The final phase concentrating on the different ways in which a user can use and exploit the EG.
\end{enumerate}
The diagrammatic symbols of Figure \ref{I3} are similar as before. We now consider each phase of the methodology briefly.

The first phase (see Figure \ref{I3}) is exactly the same as that of the original KR methodology (see Figure \ref{I1}). Briefly, the \textit{Language Engineer} designs a \textit{Lexical-Semantic Resource} composed of machine-processable language data, i.e., lexical-semantic hierarchies, designed by following the \textit{Genus-Differentia} guiding principle of Aristotle. These lexical-semantic hierarchies are then used by the \textit{Knowledge Engineer} in conjunction with the entity dataset schemas to generate the \textit{Lightweight Ontology}. The second phase of the methodology, i.e., the development of the ETG repository, is markedly different from that in Figure \ref{I1} and constitutes the core of the \textit{back} loop in \textit{from KR to KO and back}. It initiates with the still ongoing work (represented via dashed line in Figure \ref{I3}) on the adaptation of Ranganathan's guiding canons of both classification and cataloguing (see Section \ref{S4}) for the development of high-quality ETGs. Notice that the current effort (and the final goal) is to adapt all the relevant canons for classification (i.e., the different canons from the Idea Plane, Verbal Plane and Notational Plane) to generate taxonomically well-founded ETG hierarchies and the canons for cataloguing to mandate guidelines as to how (data) properties should be modelled to describe (conceptual) entities in such hierarchies. To that end, the methodology introduces a new role, that of an \textit{Ontology Engineer}, who will \textit{ensure} the development of the ETGs, and thereby, the ETG repository, in full conformance with the quality guidelines prescribed by the canons of Ranganathan suitably adapted. Notice that, with the support of Ranaganathan's canons of classification, the \textit{Ontology Engineer} would be able to generate ETGs not only restricted to macro-domains but also focused on \textit{fine-grained} micro domain of interests.

The third and the fourth phase of the methodology (see Figure \ref{I3}) is also the same as that of the original KR methodology (see Figure \ref{I1}). Briefly, the Data Scientist takes in input the lightweight ontology, the entity datasets and the relevant \textit{quality-enriched ETG} and suitably integrates them, in an iterative fashion, to ultimately generate an EG. Finally, the fourth phase concentrates on the ways in which the user can exploit the output EG. At this point, let us recollect the example of an advanced service of egocentric image annotation described in the fourth phase of the original KR methodology (in Section \ref{S3}). In addition to the above service, the methodological incorporation of Ranganathan's canons would also \textit{ensure}, in a major advance from the state-of-the-art, the generation of fine-grained classification (ETG) hierarchies which, e.g., as evidenced from (their lack in) computer vision literature\footnote{https://paperswithcode.com/task/fine-grained-image-classification}, is bound to deeply enrich (KR-based) dataset building and data quality in the entire spectrum of fine-grained image classification. 

Notice also, from Table \ref{T1}, how the roles within the integrated KO-enriched KR methodology functionally compares with that of the original KR and facet-analytical KO methodology. The role of the \textit{Language Engineer (1.1)}, \textit{Knowledge Engineer (2.1)} and \textit{User (3)} in the integrated methodology is functionally similar to that of their counterparts in the KR and KO methodology (which have been functionally mapped before). The key difference, however, is reflected in the role of \textit{Ontology Engineer (1.2 + 2.3)}, which, builds the ETG hierarchy adhering to the \textit{canons} (a completely new role absent in the other methodologies) and additionally specifies the data attributes within the ETG (part of the role of \textit{Data Scientist} in the KR and the role of \textit{Catalog Code Designer} in the KO methodology). The \textit{Data Scientist (2.2)} in the KO-enriched KR methodology is functionally mapped to the role of \textit{Cataloguer} in the KO and to a part of the role of \textit{Data Scientist} in the KR methodology. This difference is because the former \textit{Data Sceintist (2.2)} role only integrates the previous three outputs to generate the EG, whereas, the latter role of \textit{Data Sceintist (1.2 + 2.2)}, in addition to the above function, also specifies data attributes (performed by the new role of \textit{Ontology Engineer} in the KO-enriched KR methodology). In terms of artifacts, the KO-enriched KR methodology is, functionally, the same as that of the KR methodology, and, thereof, to the KO methodology. Finally, in terms of activities, the KO-enriched KR methodology adds two extra activty flows over and above the KR methodology, namely: (i) the first activity flow from \textit{Ranganathan} and \textit{Ongoing Work} to the \textit{Ontology Engineer} in terms of adapting guiding principles of CC and CCC, and (ii) the activity flow from the \textit{Ontology Engineer} to the \textit{ETG Repository} in terms of designing the repository.

% Observations
% In conclusion, there are a few observations with respect to the integrated KO-enriched KR methodology (Figure \ref{I3}). Firstly, notice that the methodology is completely \textit{general} in its scope and functionality as it can completely accommodate any level and form, e.g., in terms of languages, ontological models, domains and entities (including bibliographic entities). Secondly, the \textit{purpose} behind a specific application of the methodology is always provided \textit{by the user} and it is always the purpose which determines the selection, classification, interrelation and descriptions of entities modelled via the methodology. In the above two aspects, this methodology retains the advantages of the KR methodology and significantly improves upon the KO methodology. However, as depicted in Figure \ref{I3}, the integrated KO-enriched KR methodology, by adapting and integrating Ranganathan's guiding principles of model (and model-based data) quality within its methodological flow, provides a major advance to KR. In fact, the guiding principles (canons) for both classification and cataloguing decisively ensures the modelling quality of the ETG hierarchies, and thereby, of the final output KR model, i.e., the EG.

\section{Case Study}
\label{S6}
% ImageNet Contextualization
Let us now focus on how the incorporation of the guiding principles of modelling quality (i.e., the \textit{canons}) within the overall KO-enriched KR methodology can significantly improve the quality of those data which, within the data science research landscape, are organized and generated by exploiting KR models. To that end, we consider the case of ImageNet \cite{2009-imagenet} which is a \textit{benchmark} dataset for various computer vision tasks such as image annotation, object detection, etc. In fact, ImageNet is also the most prominent dataset within the data science landscape which is structured by populating the nodes of the WordNet lexical-semantic ontology with hundreds of thousands of images of different objects. While it has been heavily used to test and develop multiple computer vision models (e.g., \cite{DRL,DCCN}), a recent study-cum-experiment \cite{ECAI23} revealed a fundamental problem in the quality of the ImageNet ontology model which results in systematic flaws in its data quality (see \cite{2020-ICML}). 

% VC != LC + Systematic Design Flaws in Data 
The study established that, in ImageNet, there was a many-to-many mapping between, quoting \cite{ECAI23}, \emph{``the visual information encoded in an image and the intended semantics of the corresponding linguistic descriptions"}. In its detailed analysis, the study found that the underlying cause was the lack of an \textit{explicit} methodology by the ImageNet creators, backed by \textit{guiding principles} for ensuring conceptual model quality, to align the way visual classification hierarchies, e.g., of images, are modelled and the way in which lexical-semantic hierarchies are modelled. The study, quoting a previous study \cite{2020-ICML}, also noted that this misalignment between visual and linguistic classification led to two major categories of design flaws\footnote{Notice that the study \cite{ECAI23} also found an additional category of design flaw - \textit{Multi-Object Images}, the root cause of which, however, is object localization and not classification model quality and alignment.} systemic in ImageNet data annotation (and affecting all models developed and trained on it). Firstly, there are \textit{Mislabelled Images}, wherein, there is a complete misalignment in how an image is visually classified and annotated and how it is linguistically classified and annotated. A famous example is that of a \textit{birthday cake} labelled as an \textit{acoustic guitar} (see, \cite{ECAI23}). Secondly, there are \textit{Single-Object Images}, wherein, the misalignment between the visual classification and linguistic classification of an image is chiefly due to either \textit{visual polysemy} or \textit{linguistic polysemy} (several examples in \cite{ECAI23}). To that end, notice the fact that the root of all the aforementioned design flaws in ImageNet data is the lack of \textit{modelling quality} in classification, namely, in visual and linguistic classification of an image, and, in their alignment thereafter.

% Choice - C2 + C3 (Solution) + Canons driving the choices
In response to the above problems, the study proposed an overall image annotation methodology within which KO-enriched KR modelling is a \textit{key} component. Notice, however, the fact that KO-enriched KR modelling component in the said methodology has been adapted and tuned to the needs of image classification. To that end, the study proposed that any KR-based image classification and annotation exercise should be comprised of the following activities:
\begin{itemize}
    \item Firstly, \textit{Linguistic Classification}, involving the lexical-semantic hierarchical modelling of the space of linguistic labels used to annotate images. Here the meaning of each label within the linguistic hierarchy should be \textit{defined in terms of linguistically defined properties encoding a selected set of visual properties}, thereby, factoring in the alignment of visual and linguistic classification at the language level. For example, the definition of a guitar as being a \textit{“string instrument with six strings"}. 
    \item Secondly, \textit{Visual Classification}, involving the selection of visual properties of an object (in an image) and annotating it with an appropriate label from the linguistic hierarchy as defined via the linguistic classification. For example, for the object \textit{Guitar} in an image, the annotator would would not choose the label Guitar but, rather the property that it has six strings. The label \textit{Guitar} comes for free because of how it is defined via the same property in the linguistic classification hierarchy, thereby, eliminating any visual or linguistic ambiguity.
\end{itemize}

\noindent It is very interesting to note that the aforementioned modelling of visual and linguistic classification is completely \textit{guided} by the same set of \textit{canons of classification} proposed by Ranganathan (briefed in Section \ref{S4}) and adapted for KO-enriched KR-based visual classification. The reader is referred to the study in \cite{2021-CAOS,ISKO} for details regarding how the canons were adapted for visual classification and examples of several key modelling instances where the canons ensured the quality of alignment between visual and linguistic classification. 

% 3D Data Quality Improvement (accuracy, inter-annotator agreement, annotation cost)
Finally, the study in \cite{ECAI23} also performed three experiments to quantify the improvement, if any, brought by the dataset generated following the canons-based high quality alignment between visual and linguistic classification. To that end, the study modelled a dataset on a selected fragment of ImageNet, namely, the stringed musical instruments hierarchy and performed three separate experiments on: the \textit{accuracy} of computer vision (CV) models when trained on the dataset, the improvement in the \textit{inter-annotator agreement} with respect to the visual and linguistic classification, and, the improvements (if any) in the cost of image annotation. As detailed in \cite{ECAI23} (see Section 5: Evaluation), the study found significant improvements for all the above experiments and attributed the improvement to the enrichment in classification model quality facilitated by adherence to the \textit{canons}. 

\section{Related Work}
\label{S7}
% \begin{itemize}
%     \item RW Organization (1 paragraph)
%     \item RW on KR Models Quality (1 paragraph)
%     \item RW on KR Methodologies (1 paragraphs)
%     \item RW on DataScience/ML Data Quality (1 paragraph)
% \end{itemize}

There are three important research lines within Data and Information Science, namely, KR model quality, KR methodologies and data quality in data science ecosystem, which are significantly related to the current work. We consider each of the above areas briefly.

% Firstly, the entire spectrum of studies on the quality of KR models are directly related to the output of the KO-enriched KR methodology proposed in the current work. Secondly, the methodology proposed in the current work is directly comparable to the existing knowledge and data modelling methodologies existent as part of mainstream KR research. Finally, the increasingly researched upon area of data quality within the data science and machine learning research ecosystem is also very relevant to the current work. We consider each of the above areas briefly in the following paragraphs.

Firstly, we focus on literature analyzing the quality of \textit{classification semantics} of KR models of various kinds, e.g., ontologies, conceptual models. The research in, e.g., \cite{2005-FLAIRS,oops} advanced a checklist of modelling issues which directly affect the classification semantics of ontology models, including, amongst others, cycles in a hierarchy, creating polysemous elements, creating synonyms as classes, using different inconsistent naming criteria.  More recently, there have also been a lot of work in uncovering \emph{antipatterns} - taxonomically inadmissible hierarchical patterns - in KR models \cite{2014-OAP,2015-OAP}. Notice that modelling quality enrichment activity path based on Ranganathan's canons in our proposed KO-enriched KR methodology tackles all of the aforementioned modelling issues which chiefly impact the classification model of the hierarchy.

Secondly, let us focus on some of the prominent state-of-the-art KR methodologies. The highly cited methodology in \cite{1997-methontology} advanced a \emph{``life cycle to build ontologies based in evolving prototypes"}. The methodology proposed in \cite{OD101} offered the flexibility of choosing top-down, bottom-up or middle-out approaches while modelling ontologies without any explicit emphasis on modelling quality. Recently, the NeOn methodology \cite{NeOn} proposed reuse, re-engineering and merging of ontological resources and the XD methodology \cite{XD} stressed on reuse of ontology design patterns. Notice that modelling quality, especially for the latter, is crucial for integrating the various ontology patterns reused. To that end, differently from all above KR methodologies, the KO-enriched KR methodology includes a dedicated activity path for ensuring the classification modelling quality of the KR model.

Last but not the least, let us focus on data quality in data science research. The research reported in \cite{torralba2011unbiased} is an early effort in analyzing the issue of data quality. Recent research, e.g., \cite{2021-MLDatasetDev,2020-ACMFAT}, proposes adherence to best practices-based methodologies in the modelling of datasets that are attentive to limitations \cite{KD-1995} and impact. It is also worthy to note that the crowdsourcing community has also focused extensively on the problem of data quality, see, e.g., \cite{ewerth2017machines,daniel2018quality}. Some early work have also been done towards improving the process of ensuring quality, see, e.g., \cite{demartini2021managing}. In our current work, the proposed KO-enriched KR methodology is a significant advance in the aforementioned context of the methodological incorporation of quality best practices in modelling.

\section{Conclusion}
\label{S8}
To summarize, the paper provided a non-trivial functional mapping between the components of the KR and facet-analytical KO methodologies and derived an integrated KO-enriched KR methodology which subsumes all the components of the KR methodology with the major addition of the KO-based canons.

\section*{Acknowledgement}
The research has received funding from JIDEP under grant number 101058732.

\bibliographystyle{splncs04}
\bibliography{iConference}

\end{document}